\documentclass{article}
 \usepackage{array,multirow,graphicx}
\PassOptionsToPackage{numbers, compress}{natbib}

 \usepackage[final]{nips}

\usepackage[utf8]{inputenc} % allow utf-8 input
\usepackage[T1]{fontenc}    % use 8-bit T1 fonts
\usepackage{hyperref}       % hyperlinks
\usepackage{url}            % simple URL typesetting
\usepackage{booktabs}       % professional-quality tables
\usepackage{amsfonts}       % blackboard math symbols
\usepackage{nicefrac}       % compact symbols for 1/2, etc.
\usepackage{microtype}      % microtypography

\usepackage{changes}
\usepackage{booktabs}
\usepackage{graphicx} 
\usepackage{xparse}

\usepackage{subfigure} 
\usepackage{cancel}

\RequirePackage{amsthm,amsmath}
\usepackage{graphicx, amssymb, mathrsfs,amsfonts,url, bm}

\newcommand{\real}{\mathbb{R}}

\newcommand{\gap}{\,\,\,\,\,\,\,\,}

\newcommand{\btheta}{\boldsymbol\theta}

\newcommand{\ba}{\bm{a}}
\newcommand{\bA}{\bm{A}}

\newcommand{\bC}{\bm{C}}

\newcommand{\bD}{\bm{D}}

\newcommand{\bI}{\bm{I}}

\newcommand{\br}{\bm{r}}

\newcommand{\bW}{\bm{W}}
\newcommand{\bx}{\bm{x}}
\newcommand{\bX}{\bm{X}}
\newcommand{\by}{\bm{y}}
\newcommand{\bY}{\bm{Y}}

\newcommand{\bZ}{\bm{Z}}

\newcommand{\bzero}{\mathbf{0}}

\newcommand{\gammadist}{\mathcal{G}}
\newcommand{\inversegammadist}{\mathcal{G}^{-1}}

\newcommand{\normal}{\mathcal{N}}
\newcommand{\truncatednormal}{\mathcal{TN}}
\newcommand{\generaltruncatednormal}{\mathcal{GTN}}

\newcommand{\gtnsng}{\mathcal{GTNSNG}}

\usepackage{algorithm} 
\usepackage{algorithmic}

\renewenvironment{thebibliography}[1]{
\begin{oldthebibliography}{#1}
\setlength{\itemsep}{0em}
\setlength{\parskip}{0.02em}
}
{
\end{oldthebibliography}
}

\title{Comparative Study of Inference Methods for Interpolative Decomposition}

\author{
  Jun Lu 
\\
  \texttt{jun.lu.locky@gmail.com} \\
  \text{June 30, 2022}
}

\begin{document}

\maketitle

\begin{abstract}
We propose a probabilistic model with automatic relevance determination (ARD) for computing interpolative decomposition (ID), that is commonly used for low-rank approximation, feature selection, and extracting hidden patterns in data, where the matrix factors are latent variables associated with each data dimension. 
%Prior densities with support on the specified subspace are used to address the constraint on the magnitude of the factored component.
The constraint on the magnitude of the factored component is addressed using prior densities with support on the designated subspace.
Bayesian inference procedure based on Gibbs sampling is employed. We evaluate the proposed models on a range of real-world datasets including CCLE $EC50$, CCLE $IC50$, Gene Body Methylation, and Promoter Methylation datasets with different dimensions and sizes, and show that the proposed Bayesian ID algorithms with ARD lead to smaller reconstructive errors even compared to vanilla Bayesian ID algorithms with fixed latent dimension set to matrix rank.
\end{abstract}

\section{Introduction}
Matrix decomposational methods such as singular value decomposition (SVD), CUR/Skeleton decomposition, factor analysis, independent component analysis (ICA), and principal component analysis (PCA) have been used extensively over the years to reveal hidden and latent structures of matrices in many scientific and engineering areas such as recommendation systems \citep{comon2009tensor, lu2021numerical, lu2022matrix}, deep learning \citep{liu2015sparse}, computer vision \citep{goel2020survey}, collaborative filtering \citep{marlin2003modeling, lim2007variational, mnih2007probabilistic, raiko2007principal, chen2009collaborative, brouwer2017prior, lu2022matrix, lu2022bayesian}, clustering and classification \citep{li2009non, wang2013non, lu2021survey}, and machine learning in general \citep{lee1999learning}.

Moreover, low-rank matrix approximations are essential in modern machine learning and data science. 
Low-rank approximation with respect to the Frobenius norm can be easily solved with
SVD. However, it is sometimes advantageous to work with a basis that consists of a subset of the columns of the observed matrix itself for many applications \citep{halko2011finding, martinsson2011randomized}. 
One such low-rank approximation is provided by the interpolative decomposition (ID). 
%Its distinguishing feature is that it reuses columns from the original matrix. This enables it to preserve matrix properties such as sparsity and nonnegativity that also help save space in memory.
The fact that it makes use of columns from the original matrix sets it apart. As a result, it may preserve matrix properties such as nonnegativity and sparsity, which also assist conserve memory.

Interpolative decomposition is widely used as a feature selection tool which distills the essence and enables handling of massive data that was initially too vast to fit into the RAM.
%Interpolative decomposition is widely used as a feature selection tool that extracts the essence and
%allows dealing with big data which is originally too large to fit into the RAM. 
%In addition, it removes the non-relevant parts of the data which consist of errors and redundant information 
Additionally, it eliminates the incorrect and unnecessary portions of the data that consist of errors and redundant information
\citep{liberty2007randomized, halko2011finding, martinsson2011randomized, ari2012probabilistic, lu2021numerical}.
In the meantime, finding the indices associated with the spanning columns is frequently valuable for the purpose of problem analysis and data interpretation, it can be very useful to identify a subset of the matrix's columns that condenses its information. 
If the columns of the observed matrix have some specific interpretations, e.g., they are transactions in a transaction dataset, then the columns of the factored matrix in ID will have the same meaning as well.
The factored matrix obtained from interpolative decomposition is also numerically stable since its maximal magnitude is limited to a certain range.

Matrix decomposition, on the other hand, can also be viewed as statistical models in which we seek the decomposition that provides the maximum marginal likelihood (MML)
for the observed data matrix \citep{ari2012probabilistic, brouwer2017prior}. 
%Probabilistic interpretations are investigated for many popular matrix factorizations in the literature
Numerous common matrix factorizations are investigated in the literature for their probabilistic meanings
such as general real-valued matrix factorization \citep{brouwer2017prior}, nonnegative matrix factorization (NMF)
\citep{brouwer2017prior, lu2022flexible}, principal component analysis \citep{tipping1999probabilistic}, and generalized to high-order tensor factorizations \citep{schmidt2009probabilistic}. While probabilistic models can easily accommodate constraints on the specific range of the factored matrix.

In this light, our attention is drawn to the Bayesian interpolative decomposition of underlying matrices.
%In this light, we focus on the Bayesian interpolative decomposition of underlying matrices.
The interpolative decomposition of observed data matrix $\bA$ can be described by $\bA=\bC\bW+\bD$, where the data matrix $\bA= [\ba_1, \ba_2, \ldots, \ba_N]\in \real^{M\times N}$ is approximately factorized into a matrix $\bC\in \real^{M\times K}$ containing $K$ \textit{basis columns} of $\bA$ and a matrix $\bW\in \real^{K\times N}$ with absolute values of the entries no greater than 1 in magnitude \footnote{A weaker construction is to assume no entry of $\bY$ has an absolute value greater than 2. See proof of the existence of the decomposition in \citet{lu2021numerical}.}; the noise is captured by matrix $\bD\in \real^{M\times N}$. 
Training such models amounts to finding the optimal rank-$K$ approximation to the observed $M\times N$ data matrix $\bA$ under some loss functions. 
Let $\br\in \{0,1\}^N$ be the \textit{state vector} with each entry indicating the type of the corresponding column, i.e., \textit{basis column} or \textit{interpolated (remaining) column}: if $r_n=1$, then the $n$-th column $\ba_n$ is a basis column; if $r_n=0$, then $\ba_n$ is interpolated using the basis columns plus some error term. Suppose further $J$ is the set of the indices of the selected basis columns, $I$ is the set of the indices of the interpolated columns such that $J \cup I =\{1,2,\ldots, N\}$, $J=J(\br)=\{n|r_n=1\}_{n=1}^N$, and $I=I(\br)=\{n|r_n=0\}_{n=1}^N$. Then $\bC$ can be described as $\bC=\bA[:,J]$ where the colon operator implies all indices. The approximation $\bA\approx \bC\bW$ can be equivalently stated that $\bA\approx\bC\bW=\bX\bY$ where $\bX\in \real^{M\times N}$ and $\bY\in \real^{N\times N}$ so that $\bX[:,J]=\bC$, $\bX[:,I] = \bzero\in \real^{M\times (N-K)}$; $\bW = \bY[J,:]$. 
We also notice that there exists an identity matrix $\bI\in \real^{K\times K}$ in $\bW$ and $\bY$:
\begin{equation}\label{equation:submatrix_bid_identity}
	\bI = \bW[:,J] = \bY[J,J].
\end{equation}
Then, finding the low-rank interpolative decomposition of $\bA\approx\bC\bW$ can be equivalently transformed into the problem of finding the $\bA\approx\bX\bY$ with state vector $\br$ recovering the submatrix $\bC$ (see Figure~\ref{fig:id-column}).
To evaluate the approximation, \textit{reconstruction error} measured by mean squared error (MSE or Frobenius norm) is minimized (assume $K$ is known):
\begin{equation}\label{equation:idbid-per-example-loss}
	\mathop{\min}_{\bW,\bZ} \,\, \frac{1}{MN}\sum_{n=1}^N \sum_{m=1}^{M} \left(a_{mn} - \bx_m^\top\by_n\right)^2,
\end{equation}
where $\bx_m$, $\by_n$ are the $m$-th row and $n$-th column of $\bX$, $\bY$ respectively.
In this paper, we approach the magnitude constraint in $\bW$ and $\bY$ by considering the Bayesian ID models as a latent factor model where we describe a fully specified graphical model for the problem and employ Bayesian inference algorithms to find the latent components. 
%In this sense, explicit magnitude constraints are not required on the latent factors, since this is naturally taken care of by the appropriate choice of prior distribution; here we use general-truncated-normal prior. 
In this sense, explicit magnitude constraints on the latent components are not necessary because the appropriate choice of prior distribution—here, the general-truncated-normal (GTN) prior—takes care of this automatically.

This paper's primary contribution is its innovative Bayesian ID approach
%The main contribution of this paper is to propose a novel Bayesian ID method 
that can decide the latent dimension $K$ automatically which is called the \textit{GBT with ARD} model. 
To further favor flexibility and insensitivity on the hyperparameter choices, we also propose the hierarchical model known as the \textit{GBTN with ARD} algorithm, which has straightforward conditional density distributions requiring only little extra computation. 
In the meantime, the method is easy to implement. 
%We show that the proposed methods can be successfully applied to both dense and sparse datasets.
We demonstrate that both dense and sparse datasets may be successfully processed using our method.

\section{Related Work}

\subsection{Probability Distributions for Bayesian ID}\label{section:probability_distribution}
In this section, we introduce all notations and probability distributions that will be used in Bayesian ID models.

$\normal(x|\mu, \tau^{-1}) =\sqrt{\frac{ \tau}{2\pi}}\exp\{-\frac{\tau}{2} (x-\mu)^2\}$ is a normal (or Gaussian) distribution with parameters mean $\mu$ and precision $\tau$ (where variance $\sigma^2=\tau^{-1}$).

$\gammadist(x|\alpha, \beta)= \frac{\beta^\alpha}{\Gamma(\alpha)} x^{\alpha-1}\exp\{-\beta x\}u(x)$ is a Gamma distribution where $\Gamma(\cdot)$ is the gamma function and $u(x)$ is the unit step function that has a value of $1$ when $x\geq0$ and 0 otherwise.

$\inversegammadist(x|\alpha, \beta)= \frac{\beta^\alpha}{\Gamma(\alpha)} x^{-\alpha-1}\exp\{-\frac{\beta}{x}\}u(x)$ is an inverse-Gamma distribution.

$\truncatednormal(x|\mu,\tau^{-1}) =\frac{\sqrt{\frac{\tau}{2\pi}} \exp\{-\frac{\tau}{2} (x-\mu)^2 \} } 
{1-\Phi(-\mu\sqrt{\tau})} u(x)$
is a truncated-normal (TN) with zero density value below $x=0$ and renormalized to integrate to one. Parameters $\mu$ and $\tau$ are known as the ``parent mean" and ``parent precision" of the normal distribution. And $\Phi(\cdot)$ function is the cumulative distribution function of standard normal density $\normal(0,1)$.

%$\rectifieddist(x| \mu, \tau^{-1}, \lambda) =\frac{1}{C} \cdot \normal(x|\mu, \tau^{-1})\cdot \exponential(x|\lambda) =\truncatednormal(x| \frac{\tau\mu-\lambda}{\tau} , \tau^{-1}) $ is known as the rectified-normal (RN) distribution with ``parent mean" $\mu$ and ``parent precision" $\tau$. The RN distribution is a more flexible distribution with zero density below $x=0$ than the TN distribution in the sense that it has an extra variable $\lambda$ to control the behavior of the density. $C$ is a constant given by 
%$$
%C(\mu, \tau, \lambda) = \lambda\left( 1-\Phi(-\frac{\tau\mu-\lambda}{\sqrt{\tau}}) \right) \exp\left\{-\mu\lambda+\frac{\lambda^2}{2\tau}\right\},
%$$
%which is a function with respect to $\{\mu, \tau, \lambda\}$.
%Note in some texts, the TN distribution is termed as a RN distribution (e.g., \citet{schmidt2009probabilistic}). However, we here differentiate the two distributions where the reason will be clear in the sequel.

$\generaltruncatednormal(x|\mu, \frac{1}{\tau}, a, b)=\frac{\sqrt{\frac{\tau}{2\pi}} \exp \{-\frac{\tau}{2}(x-\mu)^2  \}  }{\Phi((b-\mu)\cdot \sqrt{\tau})-\Phi((a-\mu)\cdot \sqrt{\tau})}$$u(x|a,b)$
is a general-truncated-normal (GTN) with zero density below $x=a$ or above $x=b$ and renormalized to integrate to one where $u(x|a,b)$ is a step function that has a value of 1 when $a\leq x\leq b$ and 0 otherwise. Similarly, parameters $\mu$ and $\tau$ are known as the ``parent mean" and ``parent precision" of the original normal distribution. When $a=0$ and $b=\infty$, the GTN distribution reduces to a TN density.

\begin{figure*}[h]
\centering  
\vspace{-0.35cm} 
\subfigtopskip=2pt 
\subfigbottomskip=9pt 
\subfigcapskip=-5pt 
\includegraphics[width=0.95\textwidth]{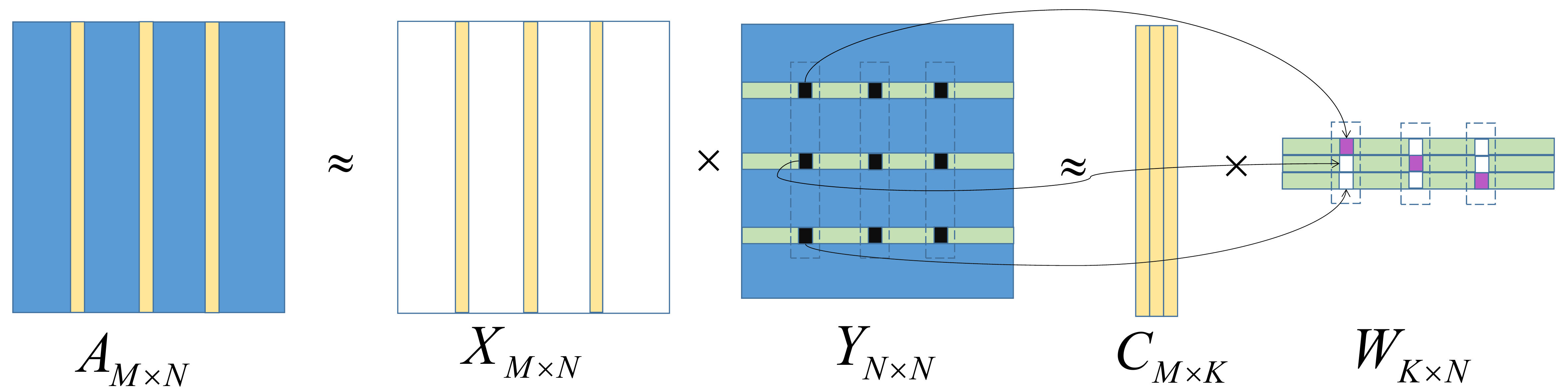}
\caption{A conceptual overview of the interpolative decomposition of matrix $\bA\in\real^{M\times N}$ where the yellow vectors denote
the basis columns of matrix $\bA$, white entries denote zero values, purple entries denote
one, blue and black entries denote elements that are not necessarily zero. The Bayesian ID models find the approximation $\bA\approx\bX\bY$ and the post processing procedure finds the approximation $\bA\approx\bC\bW$.}
\label{fig:id-column}
\end{figure*}

\subsection{Bayesian GBT and GBTN Models for Interpolative Decomposition}

\begin{figure}[h]
\centering  
\vspace{-0.55cm} 
\subfigtopskip=2pt 
\subfigbottomskip=6pt 
\subfigcapskip=-2pt 
\subfigure[GBT.]{\includegraphics[width=0.271\textwidth]{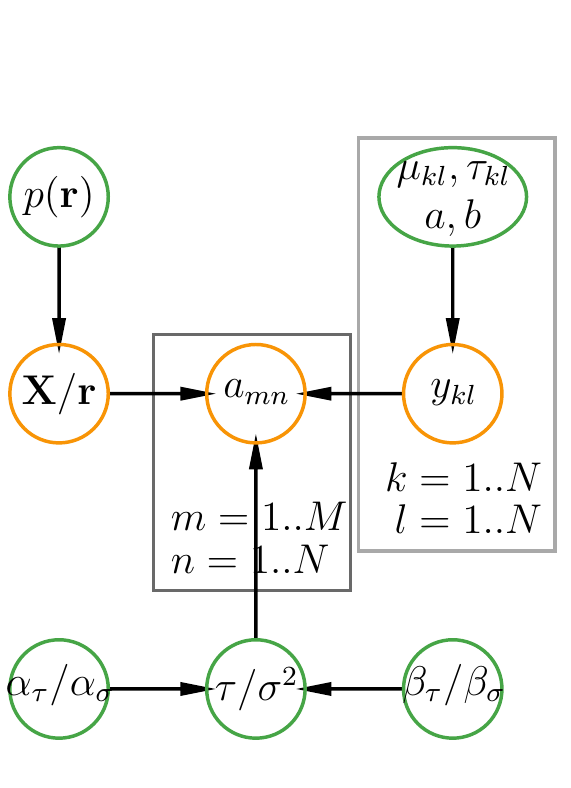} \label{fig:bmf_bid_GBT}}
\hspace{2em}
\subfigure[GBTN.]{\includegraphics[width=0.271\textwidth]{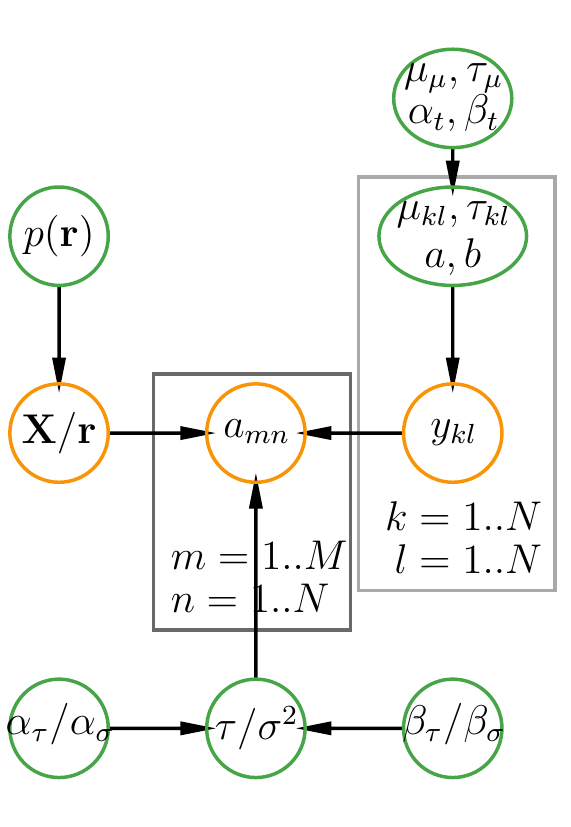} \label{fig:bmf_bid_GBTN}}
\caption{GBT and GBTN models where green circles denote prior variables, orange circles represent observed and latent variables, and plates represent repeated variables. Comma ``," in the cycles represents ``and", and ``/" in the cycles represents ``or". Parameters $a,b$ are fixed with $a=-1,b=1$ in our case; while a weaker construction can set them to $a=-2,b=2$.}
\label{fig:bmf_bids}
\end{figure}

We think of the data matrix $\bA$ as being generated according to the probabilistic generative process (see Figure~\ref{fig:bmf_bids} and also \citet{lu2022bayesian}). The observed $(m,n)$-th element $a_{mn}$ of matrix $\bA$ is modeled via a Gaussian likelihood function with variance $\sigma^2$ and mean $\bx_m^\top\by_n$ (Eq.~\eqref{equation:idbid-per-example-loss}),
\begin{equation}\label{equation:grrn_data_entry_likelihood}
	p(a_{mn} | \bx_m^\top\by_n, \sigma^2) = \normal(a_{mn}|\bx_m^\top\by_n, \sigma^2).
\end{equation}
Then, we place a conjugate prior over the data variance, i.e., an inverse-Gamma distribution with shape parameter $\alpha_\sigma$ and scale parameter $\beta_\sigma$, 
\begin{equation}\label{equation:prior_grrn_gamma_on_variance}
	p(\sigma^2 | \alpha_\sigma, \beta_\sigma) = \inversegammadist(\sigma^2 | \alpha_\sigma, \beta_\sigma).
\end{equation}
%While it can also be equivalently placed a conjugate Gamma prior over the precision and we shall not repeat the details.

We treat the latent variables $y_{kl}$'s (with $k,l\in \{1,2,\ldots,N\}$, see Figure~\ref{fig:bmf_bids}) as random variables. 
%And we need prior densities over these latent variables to express beliefs for their values, e.g., constraint with magnitude smaller than 1 in this context though there are many other constraints 
And in order to express beliefs about the values of these latent variables, we need prior densities over them, for example, a constraint with magnitude smaller than 1, even when there are many additional constraints
(e.g., semi-nonnegativity in \citet{ding2008convex}, nonnegativity in \citet{lu2022flexible}, or discreteness in \citet{gopalan2014bayesian, gopalan2015scalable}).
Here we assume further that the latent variable $y_{kl}$'s are independently drawn from a GTN prior:
\begin{equation}\label{equation:rn_prior_bidd}
	p(y_{kl} | \cdot ) = \generaltruncatednormal(y_{kl} | \mu_{kl}, (\tau_{kl})^{-1}, a=-1, b=1).
\end{equation}
As mentioned, this GTN prior serves to enforce the constraint on the components $\bY$ with no entry of $\bY$ having a value greater than 1 in magnitude, and is conjugate to the Gaussian likelihood. 
%The posterior density is also a general-truncated-normal distribution. 
While in a weaker construction of interpolative decomposition, the constraint on the magnitude can be loosened to 2; the prior is flexible in that the parameters can be then set to $a=-2, b=2$ accordingly. 
%In some scenarios, the two sets of latent variables can be drawn from two different rectified-normal priors, e.g., enforcing sparsity in $\bW$ while non-sparsity in $\bZ$. And we shall not consider this case as it is not the main interest of this paper. 
%The posterior density is a truncated-normal distribution that is a special rectified-normal distribution.

\paragraph{Hierarchical prior}
To favor further flexible structures, we place a joint hyperprior over the hyperparameters $\{\mu_{kl}, \tau_{kl}\}$ of GTN density in Eq.~\eqref{equation:rn_prior_bidd}, i.e., the GTN-scaled-normal-Gamma (GTNSNG) density,
\begin{equation}
	\begin{aligned}
		&\gap p(\mu_{kl}, \tau_{kl} |\cdot) 
		%= \gtnsng(\mu_{kl}, \tau_{kl}| \mu_\mu, \tau_\mu,\alpha_t, \beta_t)\\
		%&=	C(\mu_{mk}^W, \tau_{mk}^W, \lambda_{mk}^W)
		%\cdot 
		%\normal(\mu_{mk}^W| \mu_\mu, (\tau_\mu)^{-1})\\
		%&\gap \cdot \gammadist(\tau_{mk}^W | a, b)
		%\cdot \gammadist(\lambda_{mk}^W | \alpha_\lambda, \beta_\lambda);
		=\gtnsng(\mu_{kl}, \tau_{kl}| \mu_\mu, \frac{1}{\tau_\mu},\alpha_t, \beta_t)\\
		&=	\big\{\Phi((b-\mu_\mu)\cdot \sqrt{\tau_\mu})-\Phi((a-\mu_\mu)\cdot \sqrt{\tau_\mu})\big\} \cdot 
		\normal(\mu_{kl}| \mu_\mu, (\tau_\mu)^{-1}) \cdot \gammadist(\tau_{kl} | a_t, b_t).
	\end{aligned}
\end{equation}
Due to the useful scale, this prior can decouple the parameters $\mu_{kl}, \tau_{kl}$, and as a result, their posterior conditional densities are normal and Gamma respectively.
%This prior can decouple parameters $\mu_{kl}, \tau_{kl}$, and the posterior conditional densities of them are normal, and Gamma respectively due to this convenient scale.

\paragraph{Terminology} 
We name the model in the order of the likelihood distribution, the priors we place on the factored components, and the type of hierarchical priors.
For example, if Gaussian density is selected to be the likelihood function, and the two prior densities over the factored matrices are chosen to be Gamma density and Exponential density functions respectively,
then the model will be termed as Gaussian Gamma-Exponential (GAE) model. 
In some cases, we need to further place a hyperprior over the parameters of the prior density functions, for example, we put a Gamma prior over the Gamma density, then it will further be termed as a Gaussian Gamma-Exponential Gamma (GAEA) model. 
In this sense, the Bayeisan interpolative decomposition methods are described as the \textit{GBT} and \textit{GBTN} models where \textit{B} stands for Beta-Bernoulli density intrinsically.

\subsection{Gibbs Sampler}

We apply Gibbs sampling in this paper since it is very accurate when we tend to find the true posterior conditional density. 
%Other than this method, variational Bayesian inference can be an alternative way but we shall not go into the details. 
In this section, we only shortly describe the posterior conditional density. While a step-by-step derivation is provided in \citet{lu2022bayesian} or Appendix~\ref{appendix:gbt_gbtn_derivation}.

The posterior density of $y_{kl}$ is a GTN distribution. Denote all elements of $\bY$ except $y_{kl}$ as $\bY_{-kl}$, we have
\begin{equation}\label{equation:posterior_gbt_ykl}
	\begin{aligned}
		&\gap p(y_{kl} | \bA, \bX, \bY_{-kl}, \mu_{kl}, \tau_{kl}, \sigma^2) \propto  p(\bA|\bX,\bY, \sigma^2) \cdot p(y_{kl}|\mu_{kl}, \tau_{kl} )\\
		&=\prod_{i,j=1}^{M,N} \normal \left(a_{ij}| \bx_i^\top\by_j, \sigma^2 \right)
		\cdot \generaltruncatednormal(y_{kl} | \mu_{kl}, \frac{1}{\tau_{kl}},-1,1) \\
		&
		\propto \generaltruncatednormal(y_{kl}| \widetilde{\mu},( \widetilde{\tau})^{-1}, a=-1,b=1),
	\end{aligned}
\end{equation}
where $\widetilde{\tau} =\frac{\sum_{i}^{M}  x_{ik} ^2}{\sigma^2} +\tau_{kl}$ is the posterior ``parent precision" of the GTN distribution, and 
$$
\widetilde{\mu} = \bigg(\frac{1}{\sigma^2}  \sum_{i}^{M} x_{ik}  \big(a_{il}-\sum_{j\neq k}^{N}x_{ij}
y_{jl}\big)
+\textcolor{black}{\tau_{kl}\mu_{kl}}
\bigg) \big/ \widetilde{\tau}
$$
is the posterior ``parent mean" of the GTN distribution.

Given the state vector $\br=[r_1,r_2, \ldots, r_N]^\top\in \real^N$, the relation between $\br$
and the index set $J$ is simple; $J = J(\br) = \{n|r_n = 1\}_{n=1}^N$ and $I = I(\br) = \{n|r_n = 0\}_{n=1}^N$. A new value of state vector $\br$ is to select one index $j$ from index set $J$ and another index $i$ from index sets $I$ (we note that $r_j=1$ and $r_i=0$ for the old state vector $\br$) such that 
\begin{equation}\label{equation:postrerior_gbt_rvector}
	\begin{aligned}
		o_j &= 
		\frac{p(r_j=0, r_i=1|\bA,\sigma^2, \bY, \br_{-ji})}
		{p(r_j=1, r_i=0|\bA,\sigma^2, \bY, \br_{-ji})}\\
		&=
		\frac{p(r_j=0, r_i=1)}{p(r_j=1, r_i=0)} \times
		\frac{p(\bA|\sigma^2, \bY, \br_{-ji}, r_j=0, r_i=1)}{p(\bA|\sigma^2, \bY, \br_{-ji}, r_j=1, r_i=0)},
	\end{aligned}
\end{equation}
where $\br_{-ji}$ denotes all elements of $\br$ except $j$-th and $i$-th entries.
Trivially, we can set $p(r_j=0, r_i=1)=p(r_j=1, r_i=0)$. Then the full conditionally probability of $p(r_j=0, r_i=1|\bA,\sigma^2, \bY, \br_{-ji})$ can be calculated by 
\begin{equation}\label{equation:postrerior_gbt_rvector222}
	p(r_j=0, r_i=1|\bA,\sigma^2, \bY, \br_{-ji}) = \frac{o_j}{1+o_j}.
\end{equation}

Finally, by conjugacy, the conditional posterior density of $\sigma^2$ is an inverse-Gamma distribution:
\begin{equation}\label{equation:posterior_gnt_sigma2}
	\begin{aligned}
		&\gap p(\sigma^2 | \bX, \bY, \bA)
		= \inversegammadist(\sigma^2 | \widetilde{\alpha_\sigma}, \widetilde{\beta_\sigma}),
	\end{aligned}
\end{equation}
where $\widetilde{\alpha_\sigma} = \frac{MN}{2}+\alpha_\sigma$, 
$\widetilde{\beta_\sigma}=\frac{1}{2} \sum_{i,j=1}^{M,N}(a_{ij}-\bx_i^\top\by_j)^2+\beta_\sigma$.

\paragraph{Extra updates for the GBTN model}
Following the conceptual overview of the GBTN model in Figure~\ref{fig:bmf_bids}, the conditional posterior density of $\mu_{kl}$ can be obtained by
\begin{equation}\label{equation:posterior_gbt_mukl}
	\begin{aligned}
		&\gap p(\mu_{kl} | \tau_{kl}, \mu_\mu, \tau_\mu, a_t, b_t, y_{kl})\\
		&\propto \generaltruncatednormal(y_{kl} | \mu_{kl}, (\tau_{kl})^{-1}, a=-1, b=1) \cdot   \gtnsng(\mu_{kl}, \tau_{kl}| \mu_\mu, (\tau_\mu)^{-1},\alpha_t, \beta_t)\\
		&\propto \normal(\mu_{kl}| \widetilde{m},(\,\widetilde{t}\,)^{-1}),
	\end{aligned}
\end{equation}
where $\widetilde{t} = \tau_{kl}+\tau_\mu$, and $\widetilde{m} =(\tau_{kl}y_{kl}+\tau_\mu\mu_\mu)/\widetilde{t}$ are the posterior precision and mean of the normal density. Similarly, the conditional density of $\tau_{kl}$ is,
\begin{equation}\label{equation:posterior_gbt_taukl}
	\begin{aligned}
		&\gap p(\tau_{kl} | \mu_{kl}, \mu_\mu, \tau_\mu, a_t, b_t, y_{kl})\\
		&\propto \generaltruncatednormal(y_{kl} | \mu_{kl}, (\tau_{kl})^{-1}, a=-1, b=1)\cdot  \gtnsng(\mu_{kl}, \tau_{kl}| \mu_\mu, (\tau_\mu)^{-1},\alpha_t, \beta_t)\\
		&\propto \gammadist(\tau_{kl} | \widetilde{a}, \widetilde{b}),
	\end{aligned}
\end{equation}
where $\widetilde{a} = a_t+1/2$ and $\widetilde{b}=b_t +  \frac{(y_{kl}- \mu_{kl})^2}{2}$ are the posterior parameters of the Gamma density.
The full procedure is then formulated in Algorithm~\ref{alg:gbtn_gibbs_sampler}.

For the GBT and GBTN models, in all circumstances, the maximal magnitude of the factored matrix $\bW$ is no greater than 1 because the prior densities in Bayesian ID models guarantee the magnitude constraint. However, in many other methods, e.g., the randomized algorithm for ID, a weaker interpolative decomposition with a maximal magnitude of no more than 2 is found instead \citep{liberty2007randomized}.

\begin{algorithm}[!htb] 
\caption{Gibbs sampler for GBT and GBTN ID models. The procedure presented here can be inefficient but is explanatory. While a vectorized manner can be implemented to find a more efficient algorithm. By default, weak priors are $a=-1, b=1,\alpha_\sigma=0.1, \beta_\sigma=1$, ($\{\mu_{kl}\}=0, \{\tau_{kl}\}=1$) for GBT, ($\mu_\mu =0$, $\tau_\mu=0.1, \alpha_t=\beta_t=1$) for GBTN.} 
\label{alg:gbtn_gibbs_sampler}  
\begin{algorithmic}[1] 
%		\STATE {\bfseries Input:} Choose parameters $\alpha_\sigma, \beta_\sigma, \mu_\mu, \tau_\mu, a,  b, \alpha_\lambda, \beta_\lambda$;
\FOR{$t=1$ to $T$}
\STATE Sample state vector $\br$ from Eq.~\eqref{equation:postrerior_gbt_rvector222};
\STATE Update matrix $\bX$ by $\bA[:,J]$ where index vector $J$ is the index of $\br$ with value 1 and set $\bX[:,I]=\bzero$ where index vector $I$ is the index of $\br$ with value 0;
\STATE Sample $\sigma^2$ from $p(\sigma^2 | \bX,\bY, \bA)$ in Eq.~\eqref{equation:posterior_gnt_sigma2}; 
\FOR{$k=1$ to $N$} 
\FOR{$l=1$ to $N$} 
\STATE Sample $y_{kl}$ from Eq.~\eqref{equation:posterior_gbt_ykl};
\STATE (GBTN only) Sample $\mu_{kl}$ from Eq.~\eqref{equation:posterior_gbt_mukl};
\STATE (GBTN only) Sample $\tau_{kl}$ from Eq.~\eqref{equation:posterior_gbt_taukl};
\ENDFOR
\ENDFOR
\STATE Output loss in Eq.~\eqref{equation:idbid-per-example-loss}, and stop iteration if it converges.
\ENDFOR
\STATE Output averaged loss in Eq.~\eqref{equation:idbid-per-example-loss} for evaluation after burn-in iterations.
\end{algorithmic} 
\end{algorithm}

\section{Bayesian Interpolative Decomposition with Automatic Relevance Determination (ARD)}

\subsection{Automatic Relevance Determination (ARD)}
We furthermore extend the Bayesian
models with automatic relevance determination (ARD) to eliminate the need
for model selection.
Given the state vector $\br=[r_1,r_2, \ldots, r_N]^\top\in \real^N$ whose index sets are $J = J(\br) = \{n|r_n = 1\}_{n=1}^N$ and $I = I(\br) = \{n|r_n = 0\}_{n=1}^N$. A new value of state vector $\br$ is to select one index $j$ from either the index set $J$ or the index set $I$
such that 
\begin{equation}\label{equation:postrerior_gbt_rvector_ard}
	\begin{aligned}
		o_j &= 
		\frac{p(r_j=0|\bA,\sigma^2, \bY, \br_{-j})}
		{p(r_j=1|\bA,\sigma^2, \bY, \br_{-j})}=
		\frac{p(r_j=0)}{p(r_j=1)} \times
		\frac{p(\bA|\sigma^2, \bY, \br_{-j}, r_j=0)}{p(\bA|\sigma^2, \bY, \br_{-j}, r_j=1)},
	\end{aligned}
\end{equation}
where $\br_{-j}$ denotes all elements of $\br$ except $j$-th element.
Compare Eq.~\eqref{equation:postrerior_gbt_rvector_ard} with Eq.~\eqref{equation:postrerior_gbt_rvector}, we may find that in the former equation, the number of selected columns is not fixed now. Therefore, we let the inference decide the number of columns in basis matrix $\bC$ of interpolative decomposition.
Again, we can set $p(r_j=0)=p(r_j=1)=0.5$. Then the full conditionally probability of $p(r_j=0, r_i=1|\bA,\sigma^2, \bY, \br_{-ji})$ can be calculated by 
\begin{equation}\label{equation:postrerior_gbt_rvector222_ard}
	p(r_j=0|\bA,\sigma^2, \bY, \br_{-j}) = \frac{o_j}{1+o_j}.
\end{equation}
Then the full algorithm of GBT and GBTN with ARD is described in Algorithm~\ref{alg:gbtn_gibbs_sampler_withard} where the difference lies in that we need to iterate over all elements of the state vector rather than just one or two elements of it. We aware that many elements in the state vector can change their signs making the update of matrix $\bY$ (see Figure~\ref{fig:id-column}) unstable. Therefore, we also define a number of \textit{critical steps $\nu$}: after sampling the whole state vector $\br$, we update several times (here we repeat $\nu$ times) for the matrix $\bY$ and its related parameters (the difference is highlighted in blue of Algorithm~\ref{alg:gbtn_gibbs_sampler_withard}).

\begin{algorithm}[h] 
\caption{Gibbs sampler for GBT and GBTN ID with \textbf{ARD} models.  The procedure presented here can be inefficient but is explanatory. While a vectorized manner can be implemented to find a more efficient algorithm. By default, weak priors are $a=-1, b=1,\alpha_\sigma=0.1, \beta_\sigma=1$, ($\{\mu_{kl}\}=0, \{\tau_{kl}\}=1$) for GBT, ($\mu_\mu =0$, $\tau_\mu=0.1, \alpha_t=\beta_t=1$) for GBTN. \textcolor{blue}{Number of critical steps: $\nu$}.} 
\label{alg:gbtn_gibbs_sampler_withard}  
\begin{algorithmic}[1] 
\FOR{$t=1$ to $T$}
\FOR{\textcolor{blue}{$j=1$ to $N$}}
\STATE \textcolor{blue}{Sample state vector element $r_j$ from Eq.~\eqref{equation:postrerior_gbt_rvector222_ard}};
\ENDFOR
\STATE Update matrix $\bX$ by $\bA[:,J]$ where index vector $J$ is the index of $\br$ with value 1 and set $\bX[:,I]=\bzero$ where index vector $I$ is the index of $\br$ with value 0;
\STATE Sample $\sigma^2$ from $p(\sigma^2 | \bX,\bY, \bA)$ in Eq.~\eqref{equation:posterior_gnt_sigma2}; 
\FOR{\textcolor{blue}{$n=1$ to $\nu$}}
\FOR{$k=1$ to $N$} 
\FOR{$l=1$ to $N$} 
\STATE Sample $y_{kl}$ from Eq.~\eqref{equation:posterior_gbt_ykl};
\STATE (GBTN only) Sample $\mu_{kl}$ from Eq.~\eqref{equation:posterior_gbt_mukl};
\STATE (GBTN only) Sample $\tau_{kl}$ from Eq.~\eqref{equation:posterior_gbt_taukl};
\ENDFOR
\ENDFOR
\ENDFOR
\STATE Output loss in Eq.~\eqref{equation:idbid-per-example-loss}, stop iteration if it converges.
\ENDFOR
\STATE Output averaged loss in Eq.~\eqref{equation:idbid-per-example-loss} for evaluation after burn-in iterations.
\end{algorithmic} 
\end{algorithm}

\subsection{Post Processing}
The Gibbs sampling algorithm finds the approximation $\bA\approx \bX\bY$ where $\bX\in \real^{M\times N}$ and $\bY\in \real^{N\times N}$. As stated above, the redundant columns in $\bX$ and redundant rows in $\bY$ can be removed by the index vector $J$: 
$$
\begin{aligned}
	\bC=\bX[:,J]=\bA[:,J],\gap \text{and}\gap \bW= \bY[J,:].
\end{aligned}
$$
Since the submatrix $\bY[J,J]=\bW[:,J]$ (Eq.~\eqref{equation:submatrix_bid_identity}) from the Gibbs sampling procedure is not enforced to be an identity matrix (as required in the interpolative decomposition). We need to set it to be an identity matrix manually. This will basically reduce the reconstructive error further. 
The post processing procedure is shown in Figure~\ref{fig:id-column}.

%\subsection{Computational Complexity}
%The adopted Gibbs sampling method for GBT and GBTN models have complexity $\mathcal{O}(MN^3)$ where the most costs come from the update on the conditional density of $y_{kl}$. In the meantime, all the methods we use to compare the results (GEE, GTT, GTTN) have complexity $\mathcal{O}(MNK^2)$. Compared to the GTTN model, the proposed GRRN model only has an extra cost on the update of $\lambda_{mk}^W$ which does not amount to the bottleneck of the algorithm.

\begin{table}[h]
\centering
\begin{tabular}{lllll}
\hline
Dataset        & Rows & Columns & Fraction observed & Matrix rank\\ \hline
%GDSC $IC_{50}$ & 707  & 139     & 0.806         \\
CCLE $EC50$ & 502 & 48  &0.632   & 24\\
CCLE $IC50$ & 504 & 48 &0.965   & 24\\
Gene body meth. &160 & 254 &1.000   & 160\\
Promoter meth. & 160  & 254    & 1.000           &160\\
%CTRP $EC50$ &887 &1090 &0.801\\
%MovieLens 100K & 943  & 2946    & 0.072         \\ 
\hline
\end{tabular}
\caption{Overview of the CCLE $EC50$, CCLE $IC50$, Gene Body Methylation, and Promoter Methylation datasets, giving the number of rows, columns, the fraction of entries that are observed, and the matrix rank.}
\label{table:datadescription_ard}
\end{table}
\begin{figure}[h]
	\centering  
		\vspace{-0.35cm} 
	\subfigtopskip=2pt 
	\subfigbottomskip=2pt 
	\subfigcapskip=-5pt 
	\subfigure{\includegraphics[width=0.231\textwidth]{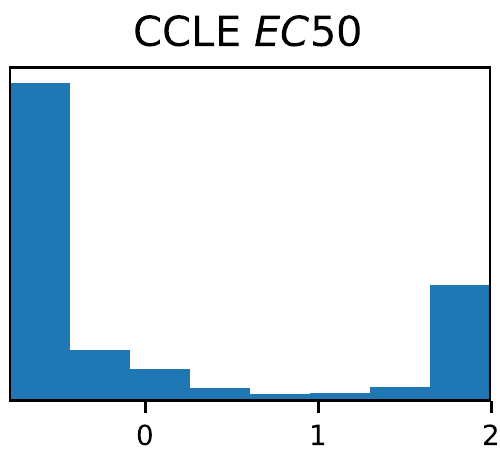} \label{fig:plot_ccle_ec}}
	\subfigure{\includegraphics[width=0.231\textwidth]{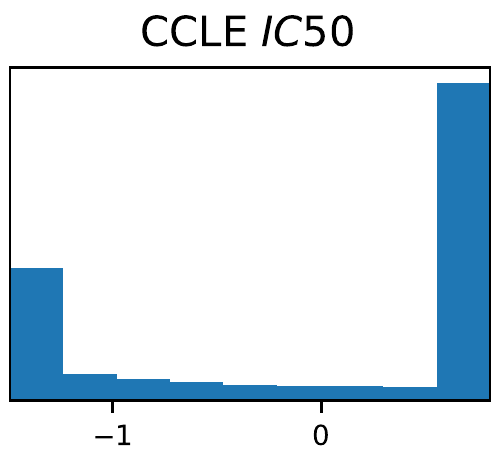} \label{fig:plot_ccle_ic}}
	\subfigure{\includegraphics[width=0.231\textwidth]{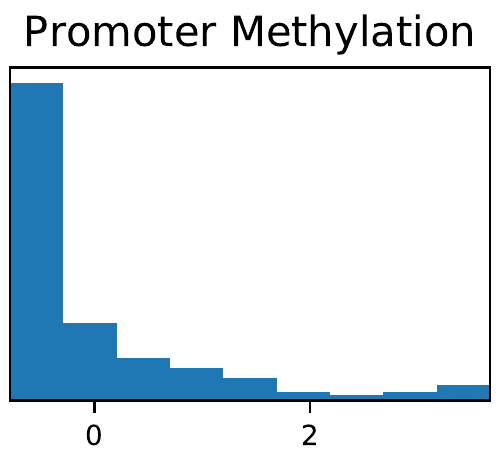} \label{fig:plot_ctrp}}
	\subfigure{\includegraphics[width=0.231\textwidth]{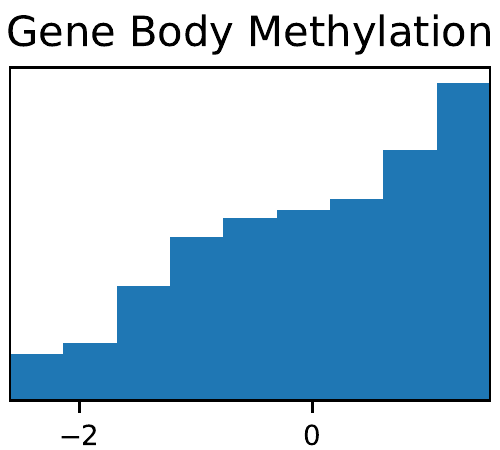} \label{fig:plot_movielens100}}
	\caption{Data distribution of CCLE $EC50$, CCLE $IC50$, Gene Body Methylation, and Promoter Methylation datasets.}
	\label{fig:datasets_bids_ard}
\end{figure}
\section{Experiments}\label{section:ader_experiments}
%RRQR \citep{cheng2005compression}, randomized ID \citep{liberty2007randomized}

To assess the proposed algorithms and highlight the key benefits of the proposed Bayesian ID with ARD methods,
%To evaluate the strategy and demonstrate the main advantages of the proposed Bayesian ID with ARD method, 
we conduct experiments with different analysis tasks; and different datasets including 
Cancer Cell Line Encyclopedia (CCLE $EC50$ and CCLE $IC50$ datasets \citep{barretina2012cancer}), 
cancer driver genes (Gene Body Methylation \citep{koboldt2012comprehensive}), and  the promoter region (Promoter Methylation \citep{koboldt2012comprehensive}) 
from bioinformatics.
Following \citet{brouwer2017prior}, we preprocess these datasets by capping high values to 100 and undoing the natural log transform for the former three datasets. Then we standardize to have zero mean and unit variance for all datasets and fill missing entries by 0. 
Finally, we copy every column twice (for the CCLE $EC50$ and CCLE $IC50$ datasets) in order to increase redundancy in the matrix; while for the latter two (Gene Body Methylation and Promoter Methylation datasets), the number of columns is already larger than the matrix rank such that we do not increase any redundancy. 
A summary of the four datasets is reported in Table~\ref{table:datadescription_ard} and their distributions are presented in Figure~\ref{fig:datasets_bids_ard}.

In all scenarios, the same parameter initialization is adopted when conducting different tasks. 
Experimental evidence reveals that post processing procedure can increase performance to a minor level, and that the outcomes of the GBT and GBTN models are relatively similar.
For clarification, we only provide the findings of the GBT model after post processing.
We compare the results of ARD versions of GBT and GBTN with vanilla GBT and GBTN algorithms. 
In a wide range of experiments across various datasets, the GBT or GBTN with ARD models improve reconstructive error and leads to performance that is as good or better than the vanilla GBT or GBTN methods in low-rank ID approximation.

In order to measure
overall decomposition performance, we use mean squared error (MSE, Eq.~\eqref{equation:idbid-per-example-loss}), which measures the similarity between the true and reconstructive; the smaller the better performance.

\begin{figure*}[h]
\centering  
\vspace{-0.15cm} 
\subfigtopskip=2pt 
\subfigbottomskip=0pt 
\subfigcapskip=-2pt 
\subfigure[Convergence of the models on the 
CCLE $EC50$, CCLE $IC50$, Gene Body Methylation, and Promoter Methylation datasets, measured by the data fit (MSE). The algorithm almost converges in less than 50 iterations.]{\includegraphics[width=1\textwidth]{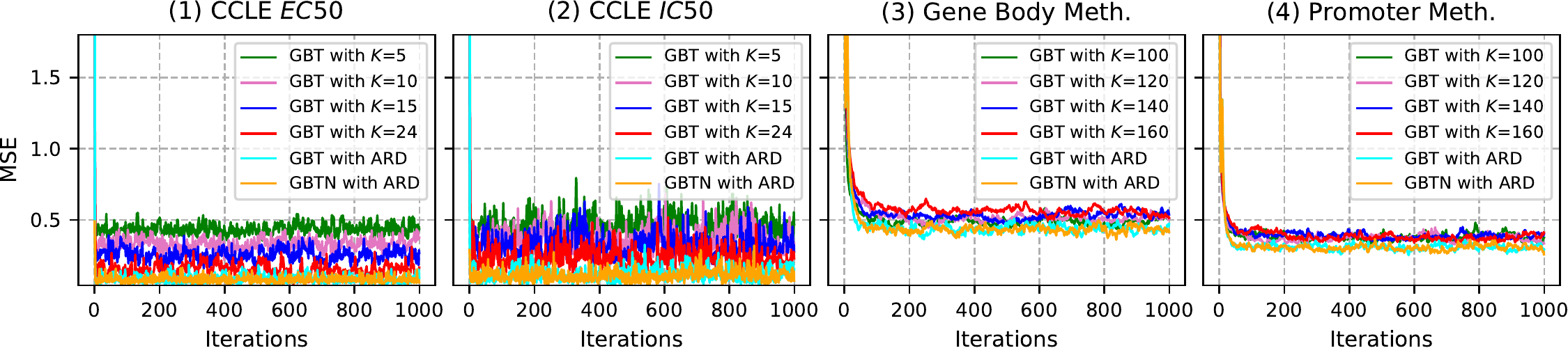} \label{fig:convergences_BIDs_ard}}
\subfigure[Averaged autocorrelation coefficients of samples of $y_{kl}$ computed using Gibbs sampling on the 
CCLE $EC50$, CCLE $IC50$, Gene Body Methylation, and Promoter Methylation datasets.]{\includegraphics[width=1\textwidth]{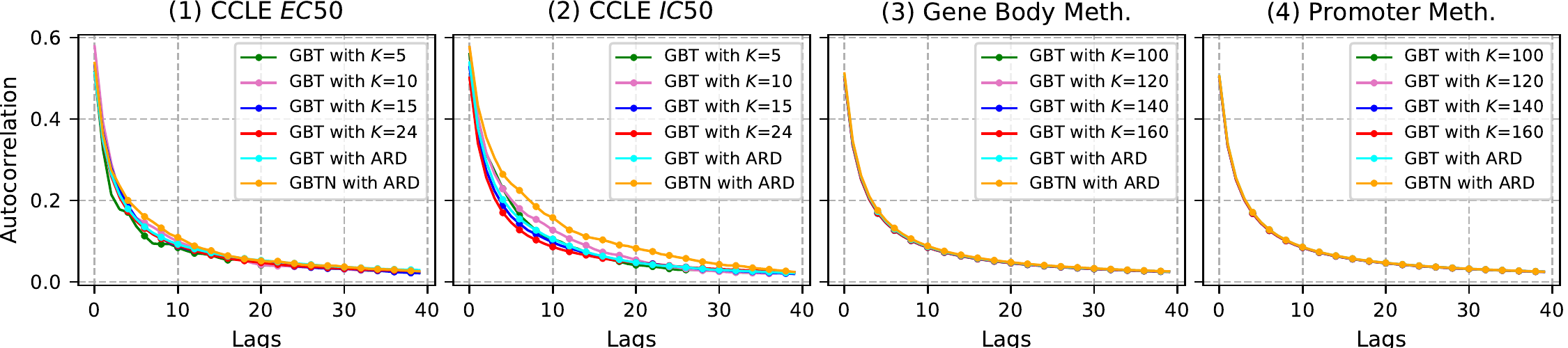} \label{fig:convergences_BIDs_autocorr_ard}}
\subfigure[Convergence of the number of selected columns on the 
CCLE $EC50$, CCLE $IC50$, Gene Body Methylation, and Promoter Methylation datasets, measured by the data fit (MSE). The algorithm almost converges in less than 100 iterations.]{\includegraphics[width=1\textwidth]{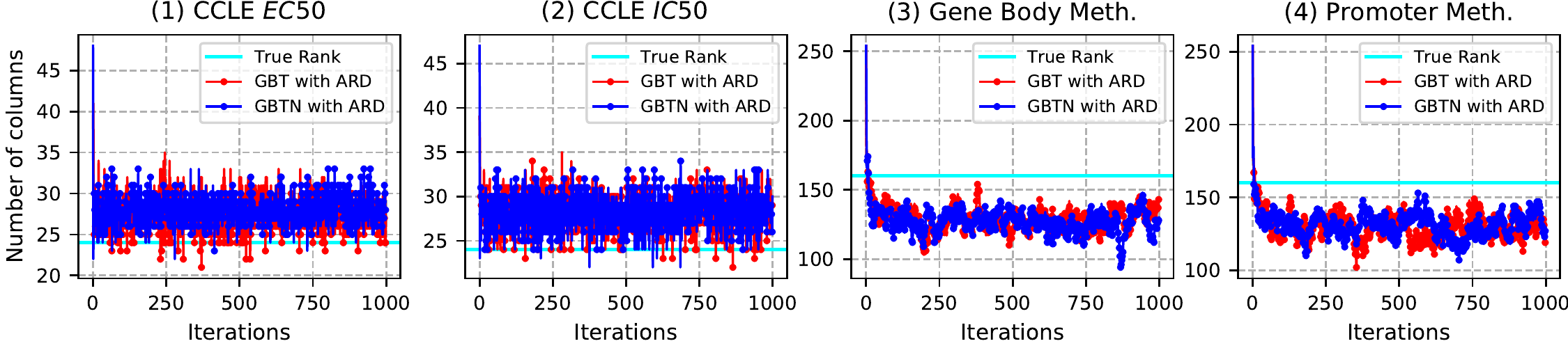} \label{fig:convergences_BIDs_numrvector_ard}}
\caption{Convergence results (upper), sampling mixing analysis (middle), and reconstructive results (lower) on the CCLE $EC50$, CCLE $IC50$, Gene Body Methylation, and Promoter Methylation datasets for various latent dimensions.
}
\label{fig:allresults_bids_ard}
\end{figure*}
\subsection{Hyperparameters}
In this experiments, we use  $a=-1, b=1,\alpha_\sigma=0.1, \beta_\sigma=1$, ($\{\mu_{kl}\}=0, \{\tau_{kl}\}=1$) for GBT, ($\mu_\mu =0$, $\tau_\mu=0.1, \alpha_t=\beta_t=1$) for GBTN, and critical steps $\nu=5$ for GBT and GBTN.
The adopted parameters are very uninformative and weak prior choices and the models are insensitive to them.
The observed or unobserved variables are initialized from random draws as long as these hyperparameters are fixed since this initialization method provides a better initial guess of the correct patterns in the matrices.
%As long as these hyperparameters are fixed, the observed or unobserved variables are initialized from random draws as this initialization procedure provides a better initial guess of the right patterns in the matrices.
In all scenarios, we run the Gibbs sampling 1,000 iterations with a burn-in of 100 iterations and a thinning of 5 iterations as the convergence analysis shows the algorithm can converge in less than 100 iterations.

\begin{table}[]
\centering
\begin{tabular}{lllllll}
\hline
& $K_1$ & $K_2$ & $K_3$ & $K_4$ & GBT (ARD) & GBTN (ARD) \\ \hline
CCLE $EC50$    &   0.354  &   0.218  & 0.131  &  0.046  & \textbf{ 0.034 } & \textbf{ 0.031 } \\
CCLE $IC50$   &   0.301  &   0.231  & 0.161  &  0.103  & \textbf{ 0.035 } & \textbf{ 0.031 } \\
Gene Body Methylation   &   0.433  &   0.443  & 0.466  &  0.492  & \textbf{ 0.363 } & \textbf{ 0.372 } \\
Promoter Methylation   &   0.323  &   0.319  & 0.350  &  0.337  & \textbf{ 0.252 } & \textbf{ 0.263 } \\
\hline
\end{tabular}
\caption{Mean squared error measure with various latent dimension $K$ parameters for CCLE $EC50$, CCLE $IC50$, Gene Body Methylation, and Promoter Methylation datasets. In all cases, $K_4$ is the full rank of each matrix, $K_1=5, K_2=10, K_3=15$ for the former two datasets, and $K_1=100, K_2=120, K_3=140$ for the latter two datasets. Results of GBT and GBTN with ARD surpass those of the GBT and GBTN with full rank $K_4$.}
\label{table:covnergence_mse_reporte_ard}
\end{table}

\subsection{Convergence and Comparative Analysis}
We first show the rate of convergence over iterations on the CCLE $EC50$, CCLE $IC50$, Gene Body Methylation, and Promoter Methylation datasets. We run GBT model with $K=5, 10, 15, 24$ for the CCLE $EC50$ and CCLE $IC50$ datasets where $K=24$ is the full rank of the matrices\footnote{The results of GBT and GBTN without ARD are close so here we only provide the results of GBT for clarity.}, 
and $K=100, 120, 140, 160$ for the Gene Body Methylation and Promoter Methylation datasets where $K=160$ is the full rank of the matrices; and the error is measured by MSE.
Figure~\ref{fig:convergences_BIDs_ard} shows the rate of convergence over iterations. Figure~\ref{fig:convergences_BIDs_autocorr_ard} shows autocorrelation coefficients of samples computed using Gibbs sampling. We observe that the mixing of the GBT and GBTN with ARD are close to those without ARD. The coefficients are less than 0.1 when the lags are more than 10 showing the mixing of the Gibbs sampler is good. 
In all experiments, the algorithm converges in less than 50 iterations.
On the CCLE $EC50$, Gene Body Methylation, and Promoter Methylation datasets, the sampling is less noisy; while on the CCLE $IC50$ dataset, the sampling of GBT without ARD seems to be noisier that those of GBT and GBTN with ARD.

Comparative results for the proposed GBT and GBTN with ARD and those without ARD on the four datasets are again shown in Figure~\ref{fig:convergences_BIDs_ard} and Table~\ref{table:covnergence_mse_reporte_ard}. 
%We also observe that, in all experiments, the MSEs of the observed entries are worse than those of all the entries.
In all experiments, the proposed GBT and GBTN with ARD achieve the smallest MSE, even compared to the non-ARD versions with latent dimension $K$ setting to full matrix rank ($K=24$ for the CCLE $EC50$ and CCLE $IC50$ datasets, and $K=160$ for the Gene Body Methylation and Promoter Methylation datasets).
Figure~\ref{fig:convergences_BIDs_numrvector_ard} shows the convergence of the number of selected columns for GBT and GBTN with ARD models on each dataset.
We observe that the samples are walking around 27 for the CCLE $EC50$ and CCLE $IC50$ datasets; and around 130 for the Gene Body Methylation and Promoter Methylation datasets. These samples are close to the true rank of each matrix.

\section{Conclusion}
The paper aims to solve the annoying choice of latent dimension issue of the Bayesian algorithms in computing the low-rank ID approximation.
We propose a computationally efficient, yet simple algorithm that needs little extra computation, that is easy to implement for interpolative decomposition, and that automatically finds the latent dimension. Overall, we show that the presented GBT and GBTN models with ARD are versatile algorithms that have good convergence results and better reconstructive performances on both sparse and dense datasets. 
Similar to vanilla GBT and GBTN models, GBT and GBTN with ARD methods are able to force the magnitude of the factored matrix to be no greater than 1 such that numerical stability is guaranteed.

%  \footnotesize
\bibliography{bib}
\bibliographystyle{sty}

\newpage
\appendix

\normalsize
\section{Derivation of Gibbs Sampler for Bayesian Interpolative Decomposition}\label{appendix:gbt_gbtn_derivation}
We provide a detailed derivation of the Gibbs sampler for the proposed Bayesian ID (with or without ARD) models in this appendix. As shown in the paper, the $(m,n)$-th entry $a_{mn}$ of the underlyuing matrix $\bA$ is modeled using a Gaussian likelihood with variance $\sigma^2$ and mean $\bx_m^\top\by_n$ (given by the latent decomposition in Eq.~\eqref{equation:idbid-per-example-loss}),
$$
\begin{aligned}
	p(a_{mn} | \bx_m^\top\by_n, \sigma^2) &= \normal(a_{mn}|\bx_m^\top\by_n, \sigma^2);\\
	p(\bA| \btheta) = \prod_{m,n=1}^{M,N}\normal \left(a_{mn}| (\bX\bY)_{mn}, \sigma^2 \right) &
	= \prod_{m,n=1}^{M,N} \normal \left(a_{mn}| (\bX\bY)_{mn}, \tau^{-1} \right),
\end{aligned}
$$
where $\btheta=\{\bX,\bY,\sigma^2\}$ denotes all parameters in the model, $\sigma^2$ is the variance, $\tau^{-1}=\sigma^2$ is the precision,

\paragraph{Prior}
We further place a conjugate prior, an inverse-Gamma distribution with shape parameter $\alpha_\sigma$ and scale parameter $\beta_\sigma$, over the variance parameter $\sigma^2$, 
$$
p(\sigma^2 | \alpha_\sigma, \beta_\sigma) = \inversegammadist(\sigma^2 | \alpha_\sigma, \beta_\sigma).
$$
Then, we assume that the latent variable $y_{kl}$'s (with $k,l\in \{1,2,\ldots,N\}$, see Figure~\ref{fig:bmf_bids}) are drawn from a GTN prior,
\begin{equation}\label{equation:rn_prior_bidd_append}
\begin{aligned}
	&\gap p(y_{kl} |\cdot) \\
	&= \generaltruncatednormal(y_{kl} | \mu_{kl}, \tau_{kl}^{-1}, a=-1, b=1)\\
	&=	 \frac{\sqrt{\frac{\tau_{kl}}{2\pi}} \exp \{-\frac{\tau_{kl}}{2}(y_{kl}-\mu_{kl})^2  \}  }{\Phi((b-\mu_{kl})\cdot \sqrt{\tau_{kl}})-\Phi((a-\mu_{kl})\cdot \sqrt{\tau_{kl}})} \cdot u(y_{kl} | a,b),
\end{aligned}
\end{equation}
where $u(x|a,b)$ is a step function that has a value of 1 when $a\leq x\leq b$ and 0 otherwise when $x<a$ or $a>b$.

\paragraph{Hyperprior}
To further favor flexible structures, we choose a joint hyperprior density over the latent parameters $\{\mu_{kl}, \tau_{kl}\}$ of GTN prior in Eq.~\eqref{equation:rn_prior_bidd_append}, i.e., the GTN-scaled-normal-Gamma (GTNSNG) prior,
\begin{equation}
	\begin{aligned}
		&\gap p(\mu_{kl}, \tau_{kl} |\cdot) \\
		&= \gtnsng(\mu_{kl}, \tau_{kl}| \mu_\mu, (\tau_\mu)^{-1},\alpha_t, \beta_t)\\
		&=	\big\{\Phi((b-\mu_\mu)\cdot \sqrt{\tau_\mu})-\Phi((a-\mu_\mu)\cdot \sqrt{\tau_\mu})\big\}
		\cdot 
		\normal(\mu_{kl}| \mu_\mu, (\tau_\mu)^{-1}) \cdot \gammadist(\tau_{kl} | a_t, b_t).
	\end{aligned}
\end{equation}
This  prior can decouple parameters $\mu_{kl}, \tau_{kl}$ due to the re-scale, and the conditional posterior densities of them are normal and Gamma distributions respectively from this convenient scale.

\paragraph{Posterior}
Following the graphical representation of the GBT (or the GBTN) model in Figure~\ref{fig:bmf_bids}, 
the conditional posterior density of $y_{kl}^W$ can be obtained by

$$
\begin{aligned}
	&\gap p(y_{kl} | \bA, \bX, \bY_{-kl}, \mu_{kl}, \tau_{kl}, \sigma^2) \propto  p(\bA|\bX,\bY, \sigma^2) \cdot p(y_{kl}|\mu_{kl}, \tau_{kl} )\\
	&=\prod_{i,j=1}^{M,N} \normal \left(a_{ij}| \bx_i^\top\by_j, \sigma^2 \right)\times
	\generaltruncatednormal(y_{kl} | \mu_{kl}, (\tau_{kl})^{-1},a=-1,b=1) \\
	&\propto
	\exp\left\{  
	-\frac{1}{2\sigma^2} \sum_{i,j=1}^{M,N} (a_{ij} - \bx_i^\top \by_j)^2
	\right\}
	\exp \{-\frac{\tau_{kl}}{2}(y_{kl}-\mu_{kl})^2  \}
	u(y_{kl} | a,b)\\
	&\propto
	\exp\left\{  
	-\frac{1}{2\sigma^2} \sum_{i}^{M} (a_{il} - \bx_i^\top \by_l)^2
	\right\}
	\exp \{-\frac{\tau}{2}(y_{kl}-\mu_{kl})^2  \}
	u(y_{kl} | a,b)\\
	%	&\propto
	%	\exp\left\{  
	%	-\frac{1}{2\sigma^2} \sum_{i}^{M} (a_{il} - x_{ik}y_{kl } - \sum_{j\neq k}^{N}x_{ij} y_{jl})^2
	%	\right\}
	%	\exp \{-\frac{\tau_{kl}}{2}(y_{kl}-\mu_{kl})^2  \}
	%	u(y_{kl} | a,b)\\
	&\propto
	\exp\left\{  
	-\frac{1}{2\sigma^2} \sum_{i}^{M} \bigg( x_{ik} ^2y_{kl }^2  + 2x_{ik} y_{kl } (\sum_{j\neq k}^{N}x_{ij} y_{jl}-a_{il})\bigg)
	\right\}
	\exp \{-\frac{\tau_{kl}}{2}(y_{kl}-\mu_{kl})^2  \}
	u(y_{kl} | a,b)\\
	&\propto
	\exp\left\{  
	-y_{kl }^2\big(\frac{\sum_{i}^{M}  x_{ik} ^2}{2\sigma^2}+\textcolor{black}{\frac{\tau_{kl}}{2}} \big)
	+y_{kl } 
	\underbrace{\bigg(\frac{1}{\sigma^2}  \sum_{i}^{M} x_{ik}  \big(a_{il}-\sum_{j\neq k}^{N}x_{ij}
		y_{jl}\big)
		+\textcolor{black}{\tau_{kl}\mu_{kl}}
		\bigg)}_{\textcolor{black}{\widetilde{\tau} \widetilde{\mu}}}
	\right\}
	%\exp \{-\frac{\tau}{2}(y_{kl}-\mu)^2  \}
	u(y_{kl} | a,b)\\
	&\propto \normal(y_{kl}| \widetilde{\mu},( \widetilde{\tau})^{-1})u(y_{kl} | a,b) 
	\propto \generaltruncatednormal(y_{kl}| \widetilde{\mu},( \widetilde{\tau})^{-1}, a=-1,b=1),
\end{aligned}
$$
where again, for simplicity, we assume the rows of $\bX$ are denoted by $\bx_i$'s and columns of $\bY$ are denoted by $\by_j$'s, $\widetilde{\tau} =\frac{\sum_{i}^{M}  x_{ik} ^2}{\sigma^2} +\tau_{kl}$ is the posterior ``parent precision" of the GTN density, and the posterior ``parent mean" of the GTN density is
$$
\widetilde{\mu} = \bigg(\frac{1}{\sigma^2}  \sum_{i}^{M} x_{ik}  \big(a_{il}-\sum_{j\neq k}^{N}x_{ij}
y_{jl}\big)
+\textcolor{black}{\tau_{kl}\mu_{kl}}
\bigg) \big/ \widetilde{\tau}.
$$

\paragraph{Update of state vector for GBT and GBTN without ARD}
Suppose further that $\br\in\{0,1\}^N$ is the state vector with each element indicating the type of the corresponding column. If $r_n=1$, then $\ba_n$ is a basis column; otherwise, $\ba_n$ is interpolated using the basis columns plus some error term. 
Given the state vector $\br=[r_1,r_2, \ldots, r_N]^\top\in \real^N$, the relation between $\br$
and the index sets $J$ is simple; $J = J(\br) = \{n|r_n = 1\}_{n=1}^N$ and $I = I(\br) = \{n|r_n = 0\}_{n=1}^N$. A new value of state vector $\br$ is to select one index $j$ from index sets $J$ and another index $i$ from index sets $I$ (we note that $r_j=1$ and $r_i=0$ for the old state vector $\br$) such that 
\begin{equation}\label{equation:postrerior_gbt_rvector_append}
o_j = 
\frac{p(r_j=0, r_i=1|\bA,\sigma^2, \bY, \br_{-ji})}
{p(r_j=1, r_i=0|\bA,\sigma^2, \bY, \br_{-ji})}
=
\frac{p(r_j=0, r_i=1)}{p(r_j=1, r_i=0)}\times
\frac{p(\bA|\sigma^2, \bY, \br_{-ji}, r_j=0, r_i=1)}{p(\bA|\sigma^2, \bY, \br_{-ji}, r_j=1, r_i=0)},
\end{equation}
where $\br_{-ji}$ denotes all elements of $\br$ except $j$-th and $i$-th entries.
Trivially, we can set $p(r_j=0, r_i=1)=p(r_j=1, r_i=0)$. Then the full conditionally probability of $p(r_j=0, r_i=1|\bA,\sigma^2, \bY, \br_{-ji})$ can be calculated by 
$$
p(r_j=0, r_i=1|\bA,\sigma^2, \bY, \br_{-ji}) = \frac{o_j}{1+o_j}.
$$

\paragraph{Update of state vector for GBT and GBTN with ARD}
For the update of the state vector in GBT  and GBTN with ARD,
a new value of state vector $\br$ is to select one index $j$ from either the index set $J$ or the index set $I$
such that 
\begin{equation}\label{equation:postrerior_gbt_rvector_ard_append}
	\begin{aligned}
		o_j &= 
		\frac{p(r_j=0|\bA,\sigma^2, \bY, \br_{-j})}
		{p(r_j=1|\bA,\sigma^2, \bY, \br_{-j})}=
		\frac{p(r_j=0)}{p(r_j=1)} \times
		\frac{p(\bA|\sigma^2, \bY, \br_{-j}, r_j=0)}{p(\bA|\sigma^2, \bY, \br_{-j}, r_j=1)},
	\end{aligned}
\end{equation}
where $\br_{-j}$ denotes all elements of $\br$ except $j$-th element.
Compare Eq.~\eqref{equation:postrerior_gbt_rvector_ard_append} with Eq.~\eqref{equation:postrerior_gbt_rvector_append}, we may find that in the former equation, the number of selected columns is not fixed now. Therefore, we let the inference decide the number of columns in basis matrix $\bC$ of interpolative decomposition.
Again, we can set $p(r_j=0)=p(r_j=1)=0.5$. Then the full conditionally probability of $p(r_j=0, r_i=1|\bA,\sigma^2, \bY, \br_{-ji})$ can be calculated by 
$$
	p(r_j=0|\bA,\sigma^2, \bY, \br_{-j}) = \frac{o_j}{1+o_j}.
$$

Finally, the conditional posterior density of $\sigma^2$ can be easily obtained by conjugacy: an inverse-Gamma distribution
$$
\begin{aligned}
	&\gap p(\sigma^2 | \bX, \bY, \bA)
	= \inversegammadist(\sigma^2 | \widetilde{\alpha_\sigma}, \widetilde{\beta_\sigma}),
\end{aligned}
$$
where $\widetilde{\alpha_\sigma} = \frac{MN}{2}+\alpha_\sigma$, 
$\widetilde{\beta_\sigma}=\frac{1}{2} \sum_{i,j=1}^{M,N}(a_{ij}-\bx_i^\top\by_j)^2+\beta_\sigma$.

\paragraph{Extra update for GBTN model}
Following the conceptual representation of the GBTN model in Figure~\ref{fig:bmf_bids}, the conditional density of $\mu_{kl}$ can be obtained by
$$
\begin{aligned}
	&\gap p(\mu_{kl} | \tau_{kl}, \mu_\mu, \tau_\mu, a_t, b_t, y_{kl})
	\propto \generaltruncatednormal(y_{kl} | \mu_{kl}, (\tau_{kl})^{-1}, a=-1, b=1)\\
&\gap\gap\gap\gap\gap\gap\gap\gap\gap \cdot \gtnsng(\mu_{kl}, \tau_{kl}| \mu_\mu, (\tau_\mu)^{-1},\alpha_t, \beta_t)\\
	&\propto\generaltruncatednormal(y_{kl} | \mu_{kl}, (\tau_{kl})^{-1}, a=-1, b=1)
	\cdot 
	\big\{\Phi((b-\mu_\mu)\cdot \sqrt{\tau_\mu})-\Phi((a-\mu_\mu)\cdot \sqrt{\tau_\mu})\big\}
 \\
&\gap\gap\gap\gap\gap\gap\gap\gap\gap		\cdot{\normal(\mu_{kl}| \mu_\mu, (\tau_\mu)^{-1})} \cdot 
	\cancel{\gammadist(\tau_{kl} | a_t, b_t)}\\
	&\propto 
	\sqrt{\tau_{kl}}\cdot \exp\left\{ -\frac{\tau_{kl}}{2} (y_{kl}-\mu_{kl})^2\right\}
	\cdot \exp\left\{ -\frac{\tau_\mu}{2}(\mu_\mu - \mu_{kl})^2  \right\}\\
	&\propto \exp\left\{  -\frac{\tau_{kl}+\tau_\mu}{2} \mu_{kl}^2 + \mu_{kl}
	\underbrace{(\tau_{kl}y_{kl}+\tau_\mu\mu_\mu)}_{\widetilde{m}\cdot \widetilde{t}}  \right\}\propto 
	\normal(\mu_{kl}| \widetilde{m},(\,\widetilde{t}\,)^{-1}),
\end{aligned}
$$
where $\widetilde{t} = \tau_{kl}+\tau_\mu$, and $\widetilde{m} =(\tau_{kl}y_{kl}+\tau_\mu\mu_\mu)/\widetilde{t}$ are the posterior precision and mean of the normal density. Similarly, the conditional density of $\tau_{kl}$ is,
$$
\begin{aligned}
	&\gap p(\tau_{kl} | \mu_{kl}, \mu_\mu, \tau_\mu, a_t, b_t, y_{kl})
	\propto \generaltruncatednormal(y_{kl} | \mu_{kl}, (\tau_{kl})^{-1}, a=-1, b=1)\\
&\gap\gap\gap\gap\gap\gap\gap\gap\gap	\cdot \gtnsng(\mu_{kl}, \tau_{kl}| \mu_\mu, (\tau_\mu)^{-1},\alpha_t, \beta_t)\\
	&\propto\generaltruncatednormal(y_{kl} | \mu_{kl}, (\tau_{kl})^{-1}, a=-1, b=1)
	\cdot 
	\big\{\Phi((b-\mu_\mu)\cdot \sqrt{\tau_\mu})-\Phi((a-\mu_\mu)\cdot \sqrt{\tau_\mu})\big\}\\
&\gap\gap\gap\gap\gap\gap\gap\gap\gap	\cdot 
	\cancel{\normal(\mu_{kl}| \mu_\mu, (\tau_\mu)^{-1})} \cdot 
	{\gammadist(\tau_{kl} | a_t, b_t)}\\
	&\propto \exp\left\{  -\tau_{kl}  \frac{(y_{kl}- \mu_{kl})^2}{2}  \right\}
	\tau_{kl}^{1/2} \tau_{kl}^{a_t-1} \exp\left\{  -b_t \tau_{kl} \right\}\\
	&\propto \exp\left\{   -\tau_{kl}\left[ b_t +  \frac{(y_{kl}- \mu_{kl})^2}{2}  \right] \right\}
	\cdot \tau_{kl}^{(a_t+1/2)-1}\\
	&\propto \gammadist(\tau_{kl} | \widetilde{a}, \widetilde{b}),
\end{aligned}
$$
where $\widetilde{a} = a_t+1/2$ and $\widetilde{b}=b_t +  \frac{(y_{kl}- \mu_{kl})^2}{2}$ are the posterior parameters of the Gamma density.

\end{document}